\documentclass[journal]{IEEEtran}
\ifCLASSINFOpdf
\else
\fi
\usepackage{cite}
\usepackage{amsmath,amssymb,amsfonts}
\usepackage{algorithmic}
\usepackage{graphicx}
\usepackage{textcomp}

\usepackage{bbding}
\usepackage{gensymb}
\usepackage{algorithm}

\usepackage{mathtools}
\usepackage{amsmath}
\usepackage{amssymb}
\usepackage{xcolor}
\usepackage{url}
\usepackage{array}
\usepackage{subfigure} 
\usepackage{stfloats}
\usepackage{multicol}

\begin{document}
%
\title{Bare Demo of IEEEtran.cls\\ for IEEE Journals}
%
%
%

\title{Synergic Adversarial Label Learning for Grading Retinal Diseases via Knowledge Distillation and Multi-task Learning}
\author{Lie Ju, Xin Wang, Xin Zhao, Huimin Lu, Dwarikanath Mahapatra, Paul Bonnington, and 
	Zongyuan Ge

\thanks{(Corresponding author: Zongyuan Ge)}
\thanks{Lie Ju and Zongyuan Ge are with Monash University, Clayton, VIC
3800 Australia, and also with Airdoc, Beijing 100000, China (E-mail:
julie334600@gmail.com, zongyuan.ge@monash.edu).
}
\thanks{Xin Wang and Xin Zhao are with Airdoc, Beijing 100000, China (E-mail:
wangxin@airdoc.com, zhaoxin@airdoc.com).}
\thanks{Huimin Lu is with Department of Mechanical and Control Engineering, Kyushu Institute of Technology, Kitakyushu 804-8550, Japan (E-mail: dr.huimin.lu@ieee.org).}
\thanks{Dwarikanath Mahapatra is with Inception Institute of Artificial Intelligence, Abu Dhabi, UAE (E-mail: dwarikanath.mahapatra@inceptioniai.org).}
\thanks{Paul Bonnington is with Monash University, Clayton, VIC
3800 Australia (E-mail: Paul.Bonnington@monash.edu).}
}

%
%

\markboth{Journal of \LaTeX\ Class Files,~Vol.~14, No.~8, August~2015}%
{Shell \MakeLowercase{\textit{et al.}}: Bare Demo of IEEEtran.cls for IEEE Journals}
%



\maketitle

\begin{abstract}
The need for comprehensive and automated screening methods for retinal image classification has long been recognized. Well-qualified doctors annotated images are very expensive and only a limited amount of data is available for various retinal diseases such as diabetic retinopathy (DR) and age-related macular degeneration (AMD).  
Some studies show that some retinal diseases such as DR and AMD share some common features like haemorrhages and exudation but most classification algorithms only train those disease models independently when the only single label for one image is available. 
Inspired by multi-task learning where additional monitoring signals from various sources is beneficial to train a robust model. We propose a method called synergic adversarial label learning (SALL) which leverages relevant retinal disease labels in both semantic and feature space as additional signals and train the model in a collaborative manner using knowledge distillation. 
Our experiments on DR and AMD fundus image classification task demonstrate that the proposed method can significantly improve the accuracy of the model for grading diseases by 5.91\% and 3.69\% respectively. 
In addition, we conduct additional experiments to show the effectiveness of SALL from the aspects of reliability and interpretability in the context of medical imaging application. 
\end{abstract}


\section{Introduction}
\label{sec:intro}
Retinal diseases such as diabetic retinopathy (DR) and age-related macular degeneration (AMD) are the leading causes of new cases of irreversible vision impairment across the world~\cite{Klein2007Overview,Gulshan2016Development,Kocur2002Visual}. It’s estimated that 25.6 million Americans with diabetes mellitus, which is almost 11\% of those people in their age group, bearing the high-risk of having DR~\cite{centers2011national,shaw2010global}. 
AMD is a leading cause of severe, irreversible vision impairment in developed countries. The prevalence of AMD varies across ethnicity and age, with increased prevalence in Caucasian and patients that are typically 50 years of age or older. Estimates suggest that the 1.75 million individuals affected by advanced AMD in at least one eye are expected to increase to nearly 3 million by year 2020~\cite{nussenblatt2007age}.  
\begin{figure*}[b]
	\centering
	\includegraphics[scale = .5]{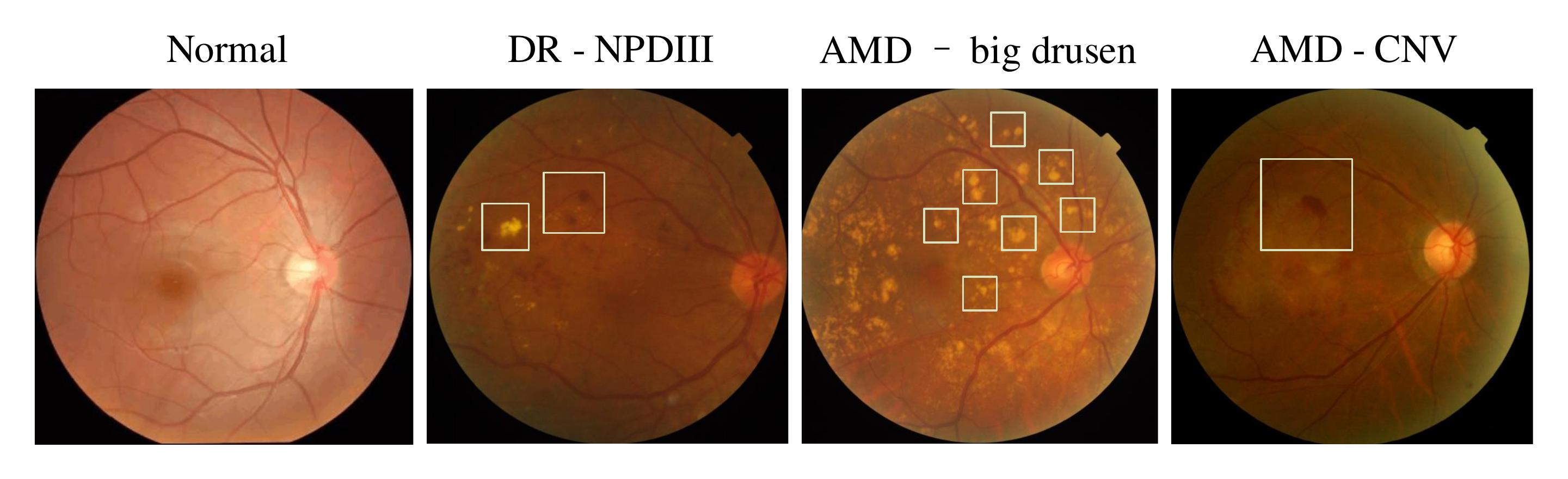}
	\caption{Some images of normal, DR and AMD. For illustration, NPDRIII, big drusen and CNV are selected as three representative levels in DR and AMD. We have marked some obvious pathological features like hemorrhagic points and drusen. Although the bright spots in DR-NPDRIII are more like exudation than big drusen, they can still be regarded as the common features in digital image space and used for our training. }
	\label{1}
\end{figure*}
At present, clinical detection of AMD and DR is a time-consuming and manual process, which requires professional doctors to screen fundus images and then give relevant reports or conclusions, usually unable to obtain reports on the same day. Such delays may often lead to miscommunication and delayed treatment. Moreover, in some areas where the incidence of the disease is high and disease detection is urgently needed, doctors with professional knowledge and necessary equipment for detection are often lacking. As the number of individuals with the disease continues to grow, the human and material resources needed to prevent blindness caused by the disease will become more inadequate. Therefore, the rapid and automatic detection of DR and AMD by computer technology is crucial for reducing the burden of ophthalmologists and providing timely morbidity analysis for a large number of patients~\cite{Yang2017Lesion}. Therefore, it is necessary to create an automatic computer-aided detection (CAD) detecting retinal diseases to make sure that the referral for treatment may be beneficial before it gets worse.
However, various fine-grained level based retinal disease conditions such as significant intra-class and inter-class variation and unbalanced training categories poses a few challenges to learn a robust CAD model.

The need for comprehensive and automated screening methods for fundus/retinal image classification has long been recognized~\cite{Faust2012Algorithms}. Taking DR as an example, we can observe how researchers analyze fundus images and design automatic screening methods. Colour features were used on Bayesian statistical classifier to classify each pixel into lesion or non-lesion classes which achieved 70\% accuracy in classifying normal retinal images as normal(\cite{wang2000effective}). \cite{sinthanayothin2002automated} used a multi-layer perceptron neural
network yielded a sensitivity of 80.21\% and a specificity of 70.66\%. \cite{acharya2008automated} used bispectral invariant
features as features for the support vector	machine classifier to classify the fundus image into normal, mild, moderate, severe and prolific DR classes which demonstrated an average accuracy of 82\% and sensitivity, specificity of 82\% and 88\% respectively.
~\cite{vujosevic2009screening} have determined single lesions to grade clinical levels of DR and diabetic macular oedema using both 1 and 3 nonmydriatic digital colour retinal images. Recently, ~\cite{seoud2015automatic, Pratt2016Convolutional} used hand-made features to represent images and random forest as the classifier. 
More recently, \cite{Gulshan2016Development} and others have tried to classify fundus images into normal and referential diabetic retinopathy (moderate and more severe) with deep convolutional neural networks (DCNN)(\cite{krizhevsky2012imagenet}) by using a large-scale retinal dataset, of which 54 U.S. licensed ophthalmologists have commented on more than 128,000 fundus images. Similarly, Sankar et al.~\cite{sankar2016earliest} used DCNN to classify lesions into normal, mild and multiple lesions. 

\begin{figure*}[!t]
	\centering
	{\includegraphics[scale=.37]{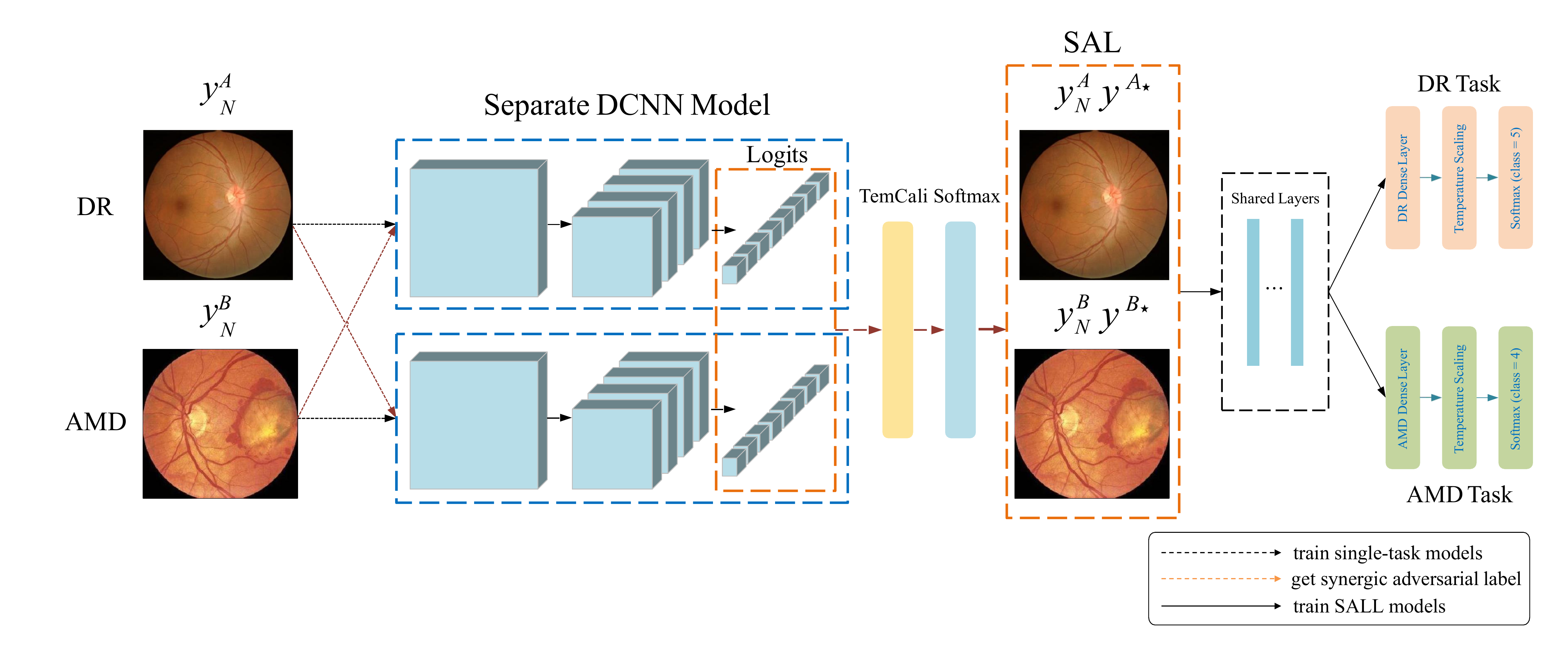}}
	\caption{An overview of our method in neural networks. In the left part of the figure, the black dotted line indicates that we use DR and AMD images to train a single-task model that only classify one specific disease. Then cross-input the image into the single-task model, and make a temperature scaling on Logits obtained from the last fully-connected layer, and finally put them into softmax, whose outputs are the synergic adversarial label. In the right part of the figure, we keep the same main network as the single-task model in the main part of the network and share weights in these layers, but modify the last fully-connected layer to two sub-tasks so that we can make better use of the synergic adversarial label. }
	\label{vvvandtree}
\end{figure*}

Although the performance of those methods is impressive in controlled experimental settings, most of them are designed and trained using one specific retinal disease only(\cite{seoud2015automatic, Pratt2016Convolutional,Gulshan2016Development,sankar2016earliest}). Some studies show that retinal diseases such as AMD and DR share some common features like hemorrhagic points and exudation as Fig.\ref{1} shows. It is such a waste not to leverage relevant retinal disease labels with common features to train the model in a collaborative manner under the scenario that good medical imaging annotations are rare and expensive. 
The collaborative label learning concept stems from knowledge distillation(\cite{Hinton2015Distilling}). \cite{Hinton2015Distilling} point out that the probability distribution of the output from a large model, it is equivalent to giving the similarity information between categories, providing additional monitoring signals. This ``smoothed" version of the signal can be used to benefit the training of a small model. We hold a hypothesis that common features among fundus/retinal images with different degrees of illness can be reflected in the logits obtained from one of the DCNN layers. Then these logits can be used as an additional signal to train the classifier.

Moreover, a DCNN model with millions of trainable parameters can easily lead to strong model overfitting(\cite{krizhevsky2012imagenet}) when training with only one category with a limited number of training images. Some previous works(\cite{Donahue2014DeCAF,mahajan2018exploring}) performed transfer learning with ImageNet pre-trained model, but fundus images are still inherently different from those natural objects as in ImageNet and CIFAR. However, some works showed that training with multiple relevant disease categories as auxiliary classes may provide regularisation effects and alleviate this issue to some extent(\cite{Chen2015Multiple,chen2014multiple}). 

To better address the problem, the categories in our training and testing set follow the standard from International Clinical Diabetic Retinopathy Disease Severity Scale~\cite{Gulshan2016Development,Haneda2010}. DR is first classified into two broad categories: nonproliferative diabetic retinopathy (NPDR) and proliferative diabetic retinopathy (PDR). NPDR is then further divided into mild, moderate, and severe stages based on progressively worsening clinical features such as microaneurysms, intraretinal haemorrhages, venous dilation, and cotton-wool spots. AMD is roughly divided into four levels, mainly according to physical signs, and corresponds to the importance of the book "Teleophthalmology"~\cite{Tele2016Xu}. (1) Early AMD which is diagnosed by medium-sized drusen within the macular area. (2) Intermediate AMD that typically have large drusen, pigment changes in the retina. (3) Late AMD with geographic atrophy which has a gradual breakdown of the light-sensitive cells in the macula. (4) Late AMD with choroidal neovascular which is abnormal blood vessels grow underneath the retina.

In this work, we present three main contributions:
\begin{enumerate}
    \item We propose a novel training method called synergic adversarial label learning (SALL). Inspired by knowledge distillation methods~\cite{Hinton2015Distilling}, our proposed method is trained with two different image classification tasks A and B (multi-task learning) whose datasets belong to two different sources but share some similarities (such as pathological similarities between DR and AMD) in the distribution of feature space. The SALL framework can be well generalized to most deep-learning-based approaches.
    \item This SALL method is able to create extra augmented data points for task A based on task B (or vice versa) by generating synergic adversarial labels to help regularise the model. The fewer data, the more significant the performance improvement of our method, which bring a lot of convenience for data collection.
    \item We conduct extensive experiments including introducing more relevant/irrelevant dataset to study the benefit of our proposed SALL approach on tasks with varying levels of similarity. The results show the superiority of SALL on several retinal diseases detection tasks.
    \item We further exploit how the proposed approach benefits the reliability and interpretability of the model, which are two important factors to consider especially in medical imaging diagnosis or automatons driving~\cite{Guo2017On}.
\end{enumerate}

\section{Related Work}

\subsection{Deep Neural Networks}
At present, in the field of computer vision, the most successful type of model to date is convolutional neural network (CNN)~\cite{krizhevsky2012imagenet}. The ImageNet Large Scale Visual Recognition Challenge (ILSVRC)~\cite{Deng2009ImageNet} has witnessed the development of CNN model in the field of computer vision. From AlexNet~\cite{krizhevsky2012imagenet}, VGGNet (\cite{simonyan2014very}), Google LeNet~\cite{Szegedy2014Going} to Residual Network (\cite{he2016deep}), the scale and depth of CNN are increasing sharply, and the recognition error rate of objects is also decreasing rapidly. At present, the recognition error rate of 1000 categories of objects in the competition is lower than that of human beings(\cite{zoph2018learning}). In addition, CNN has been successfully applied to a large number of general recognition tasks, such as object detection, semantic segmentation and contour detection ~\cite{russakovsky2015imagenet}.

\subsection{Knowledge Distillation}

The idea of knowledge distillation is firstly interpreted in~\cite{Hinton2015Distilling} to compress the knowledge in a cumbersome model (a teacher) into a smaller one(a student). The key points of knowledge distillation lie in: firstly, it generates the soft targets by distillation which raises the temperature of the final softmax until the cumbersome model produces an optimal soft set of targets; secondly, it utilizes the soft targets to train a smaller model for transfer learning. Here, the main claims about using soft targets instead of hard targets is that the soft targets provide much more information such as the distance/similarity between class through a probability distribution. Hence, the small model can be trained on much less data than the original larger ensemble model.

Since proposed, knowledge distillation has been proved to be effective in model compression with some extensions in \cite{Zagoruyko2016Paying}, \cite{Crowley2017Moonshine} and \cite{Tarvainen2018Mean}. \cite{Zagoruyko2016Paying} utilize an attention map from the teacher network to guide the student network. \cite{Tarvainen2018Mean} introduce a similar approach using mean weights. \cite{Crowley2017Moonshine} acquire the compressed model by grouping convolution channels and training the student network with an attention transfer. In general, these researches illustrate that smaller networks can be trained to perform as well as larger networks with knowledge distillation.

Knowledge distillation has been successfully applied in some other scenarios. Self-distillation is proposed to distil the knowledge from a teacher model to a student model with identical architecture. With self-distillation, the student networks realize performance than the teacher networks in \cite{Furlanello2018Born}, \cite{Bagherinezhad2018Label} and \cite{Yim2017A}. Beyond supervised learning, \cite{Lopez2015Unifying} provide theoretical and causal insight about inner workings of generalized distillation and extend it to unsupervised, semi-supervised and multitask learning scenarios. \cite{cheng2020explaining} shows that knowledge distillation (KD) usually yields more stable optimization directions than learning from raw data. Our method borrows the idea that pseudo-label of a sample can be generated by knowledge distilled from a teacher network when the true label is absent.

\subsection{Multi-Task Learning}
Multi-task learning (MTL) aims to improve learning efficiency and prediction accuracy via hard or soft parameter sharing. By sharing representations between related tasks, model generalization is improved for the knowledge learned from one task can help other tasks.  

Learning multiple related tasks simultaneously has been empirically (\cite{ando2005framework,bakker2003task,evgeniou2005learning,Evgeniou2007A,jebara2004multi,torralba2004sharing,yu2005learning,zhang2006learning}) as
well as theoretically (\cite{ando2005framework,baxter2000model,ben2003exploiting}) shown to often significantly improve performance relative to learning
each task independently. 

In MTL for computer vision, approaches often share the convolutional layers, while learning task-specific fully-connected layers. \cite{long2015learning} improve upon these models by proposing Deep Relationship Networks. Starting at the other extreme, \cite{lu2017fully} propose a bottom-up approach that starts with a thin network and dynamically widens it greedily during training using a criterion that promotes grouping of similar tasks. \cite{misra2016cross} start out with two separate model architectures just as in soft parameter sharing. They then use what they refer to as cross-stitch units to allow the model to determine in what way the task-specific networks leverage the knowledge of the other task by learning a linear combination of the output of the previous layers. All aforementioned MTL algorithms promote networks' performance via hard or soft parameter sharing. 

In this work, we use hard sharing mechanism for multi-task learning. Our method is trained with two different image tasks A and B whose datasets belong to different sources. When the available labelled training data is limited, there is an advantage in “pooling” together with data across related tasks. Furthermore, as different tasks have different noise patterns, a model that learns two tasks simultaneously is able to learn a more general representation. Learning just task A bears the risk of overfitting to task A while learning A and B jointly enables the model to obtain a better representation F through averaging the noise patterns (\cite{ruder2017overview}).

\section{Proposed Methods}
Fig.~\ref{vvvandtree} shows the overall learning structure of the SALL method we proposed in this paper. The model consisted of three steps, 1) Separate model training: train two separate DR and AMD models; 2) Generate synergic adversarial label: cross input the two categories of fundus images into the two pre-trained models from stage 1, so that each image can acquire a new label interpreting another disease; 3) Multi-task learning: we train a novel model with both synergic adversarial labels and normal labels under a multi-task learning framework to maximize the learning effectiveness;

\subsection{Separate DCNN Model Pre-training}
\label{sec: dcnn}
Notation for task A: $X_{A} = \{x_{1}^{A}, x_{2}^{A}, ..., x_{N_{1}}^{A}\}$. $Y_{A} = \{y_{1}^{A}, y_{2}^{A}, ..., y_{N_{1}}^{A}\}$. 
Notation for task B: $X_{B} = \{x_{1}^{B}, x_{2}^{B}, ..., x_{N_{2}}^{B}\}$. $Y_{B} = \{y_{1}^{B}, y_{2}^{B}, ..., y_{N_{2}}^{B}\}$. 
Model parameter trained for task A, $\theta^{A}$. 
Model parameter trained for task B, $\theta^{B}$. Our aim is to find a set of parameters $\theta^{A}$ or $\theta^{B}$ that minimizes the following cross-entropy loss
\begin{equation}\label{...}
l(\theta ) = -\frac{1}{N}[\sum_{i=1}^{N}\sum_{j=1}^{K}y_{i}\textup{log}\frac{e^{z_{j}^{(i)}}}{\sum_{n=1}^{K}e^{z_{n}^{(i)}}}]
\end{equation}
where K is the number of classes, $z=F(x,\theta)$ denotes the forward mapping of the DCNN model. There are so many algorithms can be used to optimize the $l(\theta)$ like Adaptive Moment Estimation (Adam). We obtain the last trained parameters $\theta$ thus we can get two pre-trained DCNN Models separately for DR and AMD $(DCNN-\theta^{A}, DCNN-\theta^{B})$.

\subsection{Generate Synergic Adversarial Label}
\label{sec: al}

After trained two separate models for DR and AMD, in this section, we show how to generate synergic adversarial labels.  
Let a pair of images $(x^{A}, X^{B})$ be cross input into two trained models $(DCNN-\theta^{B}, DCNN-\theta^{A})$. The synergic adversarial label can be obtained from the output of the softmax layer in the each DCNN, formally shown as follows,
\begin{equation}
y^{A\star} = F(x^{A},\theta^{B}) \qquad
y^{B\star} = F(x^{B},\theta^{A})
\end{equation}
$F$ denotes the forward mapping of the DCNN model and $y^{\star}$ represents the generated synergic adversarial label. In other words, we keep the parameters $\theta^{A}$ and $\theta^{B}$ unchanged and use the training data $X_{A} = \{x_{1}^{A}, x_{2}^{A}, ..., x_{N}^{A}\}$ as the test data, then use $DCNN-\theta^{B}$ pre-trained model to predict them. The obtained result is the synergic adversarial labels $y^{A\star}$. We take the relevant/irrelevant information between two classes as an 'adversarial' relationship between sub-tasks, which is also called \emph{negative transfer}~\cite{evgeniou2005learning} since there are both useful feature and noise shared in a multi-task learning manner (discussed in Sec. III - B).

\begin{figure}[!t]
	\centering
	\includegraphics[scale = .5]{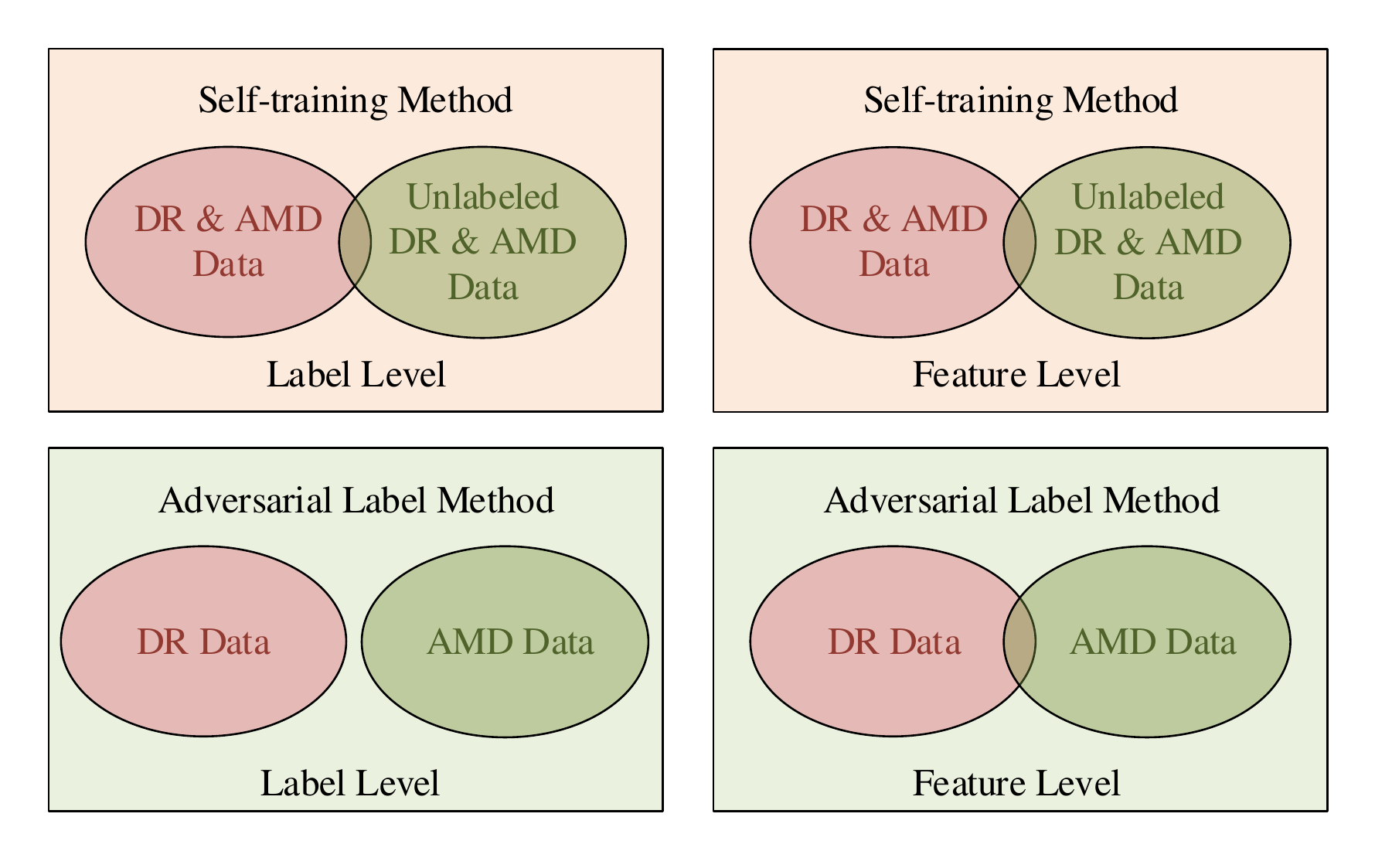}
	\caption{The difference between Self-training in Semi-supervised Learning and SALL. In Self-training, unlabeled data and labelled data not only have similarities in feature distribution but if we annotate the unlabeled data artificially, we may find that many labels in those would overlap with the existing labelled data. In our methods, if an AMD image does have both DR and AMD diseases, we can regard this as simple semi-supervised learning. But even if DR and AMD fundus images are “clean”, that is, only one disease is contained in one image, our method is still very effective.}
	\label{3}
\end{figure}

To illustrate insights from this generating process, we introduce the concept shares the similar philosophy to synergic adversarial label, called pseudo-label(~\cite{lee2013pseudo}) which is a popular training scheme being used in Semi-Supervised Learning (SSL). Like our proposed method, it first trains a model based on labelled data. Then it inputs unlabeled data into the trained model and obtains a pseudo label of unlabeled data. At last, data with pseudo-labels are added back to the data pool and retrained with original labelled data. The main difference between our synergic adversarial label and pseudo-label in SSL resides in that SSL has the assumption that those unlabelled data must belong to one of the known classes in the training set. While in our proposed method this presumption does not have to hold, see Fig.~\ref{3} for illustration. 
Pseudo-label works well in SSL because the original self-training model trained based on labelled data has provided a lot of prior knowledge for generating pseudo-label for unlabeled data since there is so much semantic overlapping in the feature space. In our problem setting, although AMD and DR are different disease classes, there are overlapping pathological evidence in feature space as discussed in Sec.~\ref{sec:intro}. Therefore, it is reasonable to use DCNN trained from AMD to implicitly interpret and leverage information from DR samples.   
For example in Fig.~\ref{1}, one DR-NPDIII sample is represented as one-hot vector $[0,0,0,1,0]$ in original DR label space. The synergic adversarial label of this sample generated from $DCNN-AMD$ is $[0.1, 0.1, 0.5, 0.3]$ for 4 different levels Normal, Small drusen, Big drusen and CNV in AMD label space. It indicates that there exist common pathological features between NPDRIII (featured by hemorrhagic points and exudation) and Big Drusen (predicting 50\% probability) as well as CNV(predicting 30\% probability). In general, we treat generated synergic adversarial labels as auxiliary or implicitly augmented data to improve training and regularisation of the model when not enough data are presented, especially in the medical imaging domain.

\subsection{Multi-task Learning}
\label{sec: ml}
\begin{figure}[!h]
	\centering
	\includegraphics[scale = .5]{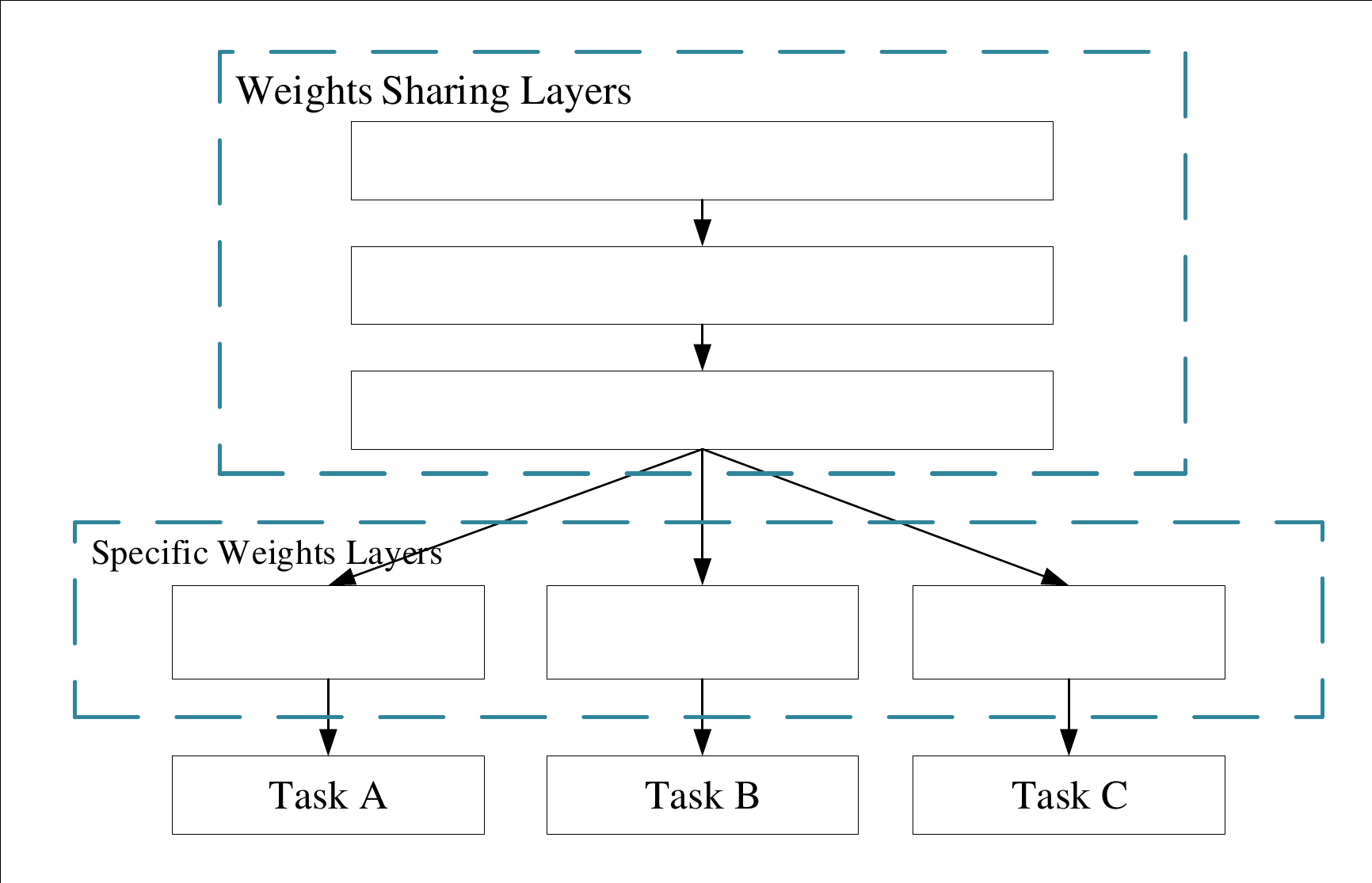}
	\caption{Hard parameters sharing for multi-task learning in deep neural networks.}
	\label{4}
\end{figure}

In this section, we introduce how synergic adversarial labels and original labels are trained jointly under the multi-task setting. We use the hard sharing mechanism for multi-task learning, as shown in the Fig.~\ref{4}. There are two main benefits of training the DR and AMD tasks in a multi-task learning manner. 1) multi-task learning effectively increases the number of training examples. Even if we do not apply the synergic adversarial label, only adding the multi-task method to training may still benefit our model, but there are more or fewer noises between them. 2) if the task noise is serious, the amount of data is small and the dimension of data is high, it is very difficult for each task to distinguish the relevant and irrelevant features. Multi-task helps to focus on the model's attention on those features that really have an impact because other tasks can provide additional evidence for the correlation and irrelevance of features. 

We set two Input Layers for DR and AMD label respectively, since for every image we have a ground truth label and a synergic adversarial label so the input is $X_{A} = \{x_{1}^{A}, x_{2}^{A}, ..., x_{N}^{A}\}$, $X_{B} = \{x_{1}^{B}, x_{2}^{B}, ..., x_{N}^{B}\}$. Compared with the separate pre-trained model, we change the last fully connected layer to two sub-fully connected layers for DR and AMD task and keep all the layers above sharing parameters. The forward computing and training parameters can be formally shown as follow:
\begin{equation}
\left\{
\begin{array}{lr}
f_{M}  =  F(X_{A},X_{B},\theta^{M})  \\
f_{A}  =  F(f_{M},\theta^{A})  \\
f_{B}  =  F(f_{M},\theta^{B})
\end{array}
\right.
\end{equation}
where $F$ is the forward computing. $\theta^{M}$ is the parameters of shared layers and $\theta^{A}$ and $\theta^{B}$ are the parameters of two sub-task layers. We use the results of the forward computing on shared layers to make an update on $\theta^{A}$ and $\theta^{B}$. For two sub-tasks, we use the same cross-entropy loss function as separate DCNN models:
\begin{equation}
l_{A}(\theta_{A}) = -\frac{1}{N}[\sum_{i=1}^{N}\sum_{j=1}^{K_{A}}y_{i}^{A}\textup{log}\frac{e^{z_{j}^{(i)}}}{\sum_{n=1}^{K_{A}}e^{z_{n}^{(i)}}}]
\end{equation}
\begin{equation}
l_{B}(\theta_{B}) = -\frac{1}{M}[\sum_{i=1}^{M}\sum_{j=1}^{K_{B}}y_{i}^{B}\textup{log}\frac{e^{z_{j}^{(i)}}}{\sum_{n=1}^{K_{B}}e^{z_{n}^{(i)}}}]
\end{equation}
for the weights sharing layers, we use the sum of the loss values of the two sublayers as the total loss:
\begin{equation}
\begin{split}
l_{total}(\theta_{M}) = l_{A}(\theta_{A}) + l_{B}(\theta_{B}) \\
\end{split}
\end{equation}
where $K_{A}$ and $K_{B}$ are the classes number of DR and AMD. We count (4) and (5) for parameters of two sub-task updated and (6) for parameters of shared layers updated as follow:

\begin{equation}  
\left\{  
\begin{array}{lr}  
\theta_{A}(t+1) = \theta_{A}(t) - \eta \cdot \Delta_{A}  \\
\theta_{B}(t+1) = \theta_{B}(t) - \eta \cdot \Delta_{B}  \\
\theta_{M}(t+1) = \theta_{M}(t) - \eta \cdot \Delta_{M} 
\end{array}  
\right.  
\end{equation}  
where $\eta$ is the learning rate. We count the $\Delta$ as:
\begin{equation}  
\left\{  
\begin{array}{lr}  
\Delta_{A} = \frac{\partial l_{A}(\theta_{A})}{\partial \theta_{A}}  \\
\Delta_{B} = \frac{\partial l_{B}(\theta_{B})}{\partial \theta_{B}}  \\
\Delta_{M} = \frac{\partial l_{total}(\theta_{M})}{\partial \theta_{M}} 
\end{array}  
\right.  
\end{equation}

The overall learning process is shown in Alg.~\ref{algorithm_1}

\begin{algorithm}[htb]
	\caption{Training multi-task network}
	\label{algorithm_1}
	\begin{algorithmic}
		\REQUIRE ~~\\
		All the images $X_{All} = \{x_{1}^{A}, x_{2}^{A}, ..., x_{N_{A}}^{A},x_{1}^{B}, x_{2}^{B}, ..., x_{N_{B}}^{B}\}$ and corresponding labels $Y = \{[y_{1}^{A},y_{1}^{A\star}], [y_{2}^{A},y_{2}^{A\star}], ..., [y_{N_{A}}^{A},y_{N_{A}}^{A\star}],$			$[y_{1}^{B\star},y_{1}^{B}], [y_{2}^{B\star},y_{2}^{B}], ..., [y_{N_{B}}^{B\star},y_{N_{B}}^{B},]\}$, initialized $\theta^{M}$, $\theta^{A}$ and $\theta^{B}$, hyper parameters $\lambda$.
		
		Note$^{*}$: shape of $y_{i}^{A}$ = shape of $y_{i}^{B\star}$ or vice versa.
		\ENSURE ~~\\
		\textbf{Step. 1:} Foward propagation:
		
		$f_{M} = F(X_{A},X_{B},\theta^{M})$
		
		$f_{A} = F(f_{M},\theta^{A})$
		
		$f_{B} = F(f_{M},\theta^{B})$
		
		\textbf{Step. 2:} Compute loss for sub-task:

		$l_{A}(\theta_{A}) = -\frac{1}{N}[\sum_{i=1}^{N}\sum_{j=1}^{K_{A}}y_{i}^{A}\textup{log}\frac{e^{z_{j}^{(i)}}}{\sum_{n=1}^{K_{A}}e^{z_{n}^{(i)}}}]$
		
		$l_{B}(\theta_{B}) = -\frac{1}{N}[\sum_{i=1}^{N}\sum_{j=1}^{K_{B}}y_{i}^{B}\textup{log}\frac{e^{z_{j}^{(i)}}}{\sum_{n=1}^{K_{B}}e^{z_{n}^{(i)}}}]$
		
		$l_{total}(\theta_{M}) = l_{A}(\theta_{A}) + l_{B}(\theta_{B})$
		
		\textbf{Step. 3:} Compute gradient:
		
		$\Delta_{A} = \frac{\partial l_{A}(\theta_{A})}{\partial \theta_{A,B}}$
		
		$\Delta_{B} = \frac{\partial l_{B}(\theta_{B})}{\partial \theta_{B}}$
		
		$\Delta_{M} = \frac{\partial l_{total}(\theta_{M})}{\partial \theta_{M}}$
		
		\textbf{Step. 4:} Update parameters:
		
		$\theta_{A}(t+1) \leftarrow  \theta_{A}(t) - \eta \cdot \Delta_{A}$
		
		$\theta_{B}(t+1) \leftarrow  \theta_{B}(t) - \eta \cdot \Delta_{B}$
		
		$\theta_{M}(t+1) \leftarrow  \theta_{M}(t) - \eta \cdot \Delta_{M}$
	\end{algorithmic}
\end{algorithm}

When applying the trained multi-task model to the classification of
a test image $X$, we can get a prediction probability distribution $P = [[p_{1}^{A},p_{2}^{A},p_{3}^{A},p_{4}^{A},p_{5}^{A}]$, $ [p_{1}^{B},p_{2}^{B},p_{3}^{B},p_{4}^{B}]]$. DR or AMD task classification can be predicted as 

\begin{equation}
Result = argmax(P,i)
\end{equation}
where $i$ denotes the dimension of $P$, and $i$=0 or 1 for different tasks. 0 for DR and 1 for AMD.

\subsection{Temperature Scaling}
\label{sec: ts}
Predicting probability estimates of the true correctness likelihood is important for classification models in many applications~\cite{Guo2017On}. It is sometimes considered that the high accuracy of the model prediction does not mean that the model is reliable, this problem is quite common in most of the deep-learning-based models(e.g.  110-layer ResNet(~\cite{he2016deep}) on CIFAR-100 dataset). Imaging that a model is unsure about one prediction but gives $99\%$ overconfident score. This will result in an unexpected outcome and prevent human assistant which may become dangerous in applications such as medical diagnosis and autonomous driving.

To solve this problem for the retinal classification problem, we add temperature scaling to two sub-tasks after the fully-connected layer. The details of temperature scaling are as follows
\begin{equation}\label{...}
\displaystyle{Outputs = \frac{e^{z_{i}/T}}{\sum_{j} e^{z_{j}/T}}}
\end{equation}
Where $T$ is the temperature we set, and $z_{i}$, which is also called logits, is the outputs of the last fully connected layer. For the SALL part, the $Outputs$ is the synergic adversarial label, which has been changed to a softer probability distribution produced by T over classes compared with before.
For the multi-task learning part, the $Outputs$ is used for back-propagation to update the $\theta^{M}$ in the multi-task networks, and the temperature scaling can help the network to produce a soft score in the hard-coding multi-task networks. When the information entering the network is some irrelevant information, it prevents the effect of irrelevant information to be over-confident and its impact on the model will be reduced. Although the feature information entering will be reduced too, we believe that the benefits of taking this risk are much higher than that of not using temperature scaling.

\section{Experiments and Discussion}

This section is organized as follows: Section IV-A gives the details of our experimental settings. Section IV-B gives a comprehensive comparative and ablation study on how each component makes our proposed methods performant. Section IV-C provides more metrics to measure the reliability of the calibrated model. Section IV-D makes an interpretability analysis through visualization. Section IV-E studies how data ratio affects the trade-off between different sub-tasks. Finally, Section IV-F further introduces a new sub-task to study the changes with varying levels of similarity between sub-tasks.

\subsection{Experimental settings}
\begin{table}[]
	\centering
	\caption{The amount of datasets.}
	\begin{tabular}{cccc}
		\hline
		
		\textbf{DR}  & \textbf{Train} & \textbf{Validation} & \textbf{Test} \\ \hline
		Normal       & 7359           & 2454                & 2454          \\
		NPDRI        & 4878           & 543                & 543          \\
		NPDRII       & 5353           & 1785                & 1785          \\
		NPDRIII      & 5040           & 840                 & 840           \\
		NPDR         & 5457           & 921                 & 921           \\ \hline
		\textbf{AMD} & \textbf{}      & \textbf{}           & \textbf{}     \\ \hline
		Normal       & 7359           & 2454                & 2454          \\
		Small Drusen & 7735           & 2579                & 2579          \\
		Big Drusen   & 5463           & 608                 & 608           \\
		CNV          & 7671           & 2558                & 2558          \\ \hline
	\end{tabular}

	\label{dataset_amount}
\end{table}

To the best of our knowledge, there are not any datasets containing both multi-label DR and AMD labels. In order to fully verify our proposed method, besides re-labelling the popular Kaggle DR~\cite{KagleDiabetic} dataset, we additional collected extra DR and AMD images from some private hospitals. All the images from either Kaggle or private hospitals are verified re-labelled by at least three ophthalmologists. The labels of one fundus image preserve only if at least two ophthalmologists are in agreement. There are no overlapping images between DR and AMD, in other words, for each image, there is only one label of DR or AMD. From the medical point of view, the probability of a person suffering from both these two diseases is very low, which is also a reasonable explanation.

All labels are set with one-hot vector style and for those who do not belong to any category (normal or healthy) are denoted as vector $[1,...0,...,0]$. 

The dataset is partitioned into training ($50\%$), validation ($25\%$) and testing ($25\%$). To increase the amount of the training set, we have used some traditional data augmentation methods as follow:

1. randomly scaled by $\pm$10\%,

2. randomly rotated by 0, 90, 180 or 360 degrees,

3. randomly flipped vertically or horizontally,

4. randomly skewed by $\pm$0.2. 

After data augmentation, the details of class distribution from the dataset are shown in Table.~\ref{dataset_amount}.

\begin{table}[]
	\centering
		\caption{Parameters of Adam}
	\begin{tabular}{p{2cm} p{2cm}}
		\hline
		\textbf{name} & \textbf{value} \\ \hline
		lr            & 1e-5           \\
		beta\_1       & 0.9            \\
		beta\_2       & 0.999          \\
		epsilon       & 1e-8           \\
		decay         & 0              \\ \hline
	\end{tabular}

	\label{adam}
\end{table}
Our methods are all implemented with Python-based on Keras with Tensorflow backend.  \cite{Gulshan2016Development} is considered to be a typical work which explored the deep learning technique on retinal diseases detection. Hence, we follow the experiment settings which leverage the InceptionV3~\cite{szegedy2016rethinking} architecture as the backbone network for a comparison study. It should be noted that we adjusted all the hyperparameters to make our baseline model perform best on our datasets. Some detailed parameters of Adam can be seen in Table.~\ref{adam}. On this basis, for DR or AMD separate network, we add a dense layer with softmax after the last FC layer, and for the multi-task network, we add two dense layers with softmax, as Fig.~\ref{vvvandtree} shows. The size of the dense layer is the level number of diseases(e.g. 5 for DR).  For a fair comparison with other baselines, all the hyper-parameters of the baseline models and our proposed methods are equally the same.

In our experiments, we train a DR or AMD disease dependent single-task model as our baseline. The performance of various methods is measured by the average accuracy rate. To reduce the performance uncertainty and bias in the training process, we train each model at least three times and make 3-fold cross-validation, then we report the average performance as our final result. 



\begin{table*}[t]
	\centering
	\small
	\caption{Experiment-I: The accuracy, confidence and error of DR and AMD. *Note: baseline model.}
	\begin{tabular}{lllllllll}
		\hline
		T & SA-label & Multi-task & DR acc           & AMD acc          & DR con           & AMD con          & DR err           & AMD err          \\ \hline
		$1^*$ & no     & no         & 75.39\%          & 79.15\%          & 85.74\%          & 86.06\%          & 65.39\%          & 78.81\%          \\
		1 & yes    & no         & 75.35\%          & 78.70\%          & 82.93\%          & 87.71\%          & 64.84\%          & 77.61\%          \\
		1 & no     & yes        & 75.91\%          & 78.81\%          & 85.06\%          & 86.85\%          & 65.72\%          & 69.27\%          \\
		1 & yes    & yes        & 76.92\%          & 79.34\%          & 84.25\%          & 85.65\%          & 63.23\%          & 67.23\%          \\
		\hline
		2 & yes    & yes        & 77.53\%          & 81.12\%          & \textbf{86.99\%} & \textbf{88.26\%} & 63.38\%          & 66.55\%          \\
		\hline
		3 & yes    & yes        & \textbf{78.01\%} & \textbf{81.25\%} & 85.68\%          & 86.15\%          & \textbf{63.10\%} & \textbf{65.75\%} \\
		3 & no     & no         & 76.65\%          & 80.13\%          & 86.07\%          & 85.07\%          & 66.02\%          & 77.11\%          \\
		3 & yes    & no         & 77.36\%          & 80.36\%          & 86.90\%          & 87.65\%          & 66.81\%          & 77.40\%          \\
		3 & no     & yes        & 77.15\%          & 79.86\%          & 85.87\%          & 86.85\%          & 65.51\%          & 70.64\%          \\ \hline
	\end{tabular}

	\label{table_1}
\end{table*}
\subsection{Comparative and Ablation Study}
For the first part of the Table.~\ref{table_1} (First upper half where temperature scaling T = 1), we present the results of baseline model w/o synergic adversarial label (Sec.~\ref{sec: al}) and w/o multi-task training (Sec.~\ref{sec: ml}). 
Compared with the baseline model trained only by single-type data (first line of the Table.~\ref{table_1}), we can find that the method with synergic adversarial labels is worse in terms of average accuracy, as shown in the second line of Table.~\ref{table_1}. It leads to the observation that if we simply add irrelevant synergic adversarial labels, most of the new training data with synergic adversarial labels are just served as noise which may confuse the model. There are not clear findings of regularisation effect for the model training with synergic adversarial labels. Although AMD and DR may share common features in some region in the retinal image, useful features are hard to be distinguished according to the adversarial signals.

Then, we evaluate how the proposed synergic adversarial label augmentation method affects the multi-task learning for retinal disease recognition. We first report the performance of the model trained with a multi-label setting without the synergic adversarial label setting.   
As shown in the 3rd row of Table.~\ref{table_1}, the accuracy of DR is slightly improved while that of AMD is slightly decreased. This shows that features learned from AMD samples benefit the DR branch training. We then combine synergic adversarial label augmentation with multi-task in our model and make an observation. Compared with the baseline model, when we set $T=3$ for temperature scaling we can observe that the accuracy of DR and AMD are improved by $2.62\%$ and $2.10\%$ respectively.
Synergic adversarial label brings extra training information under the multi-task learning setting, the sharing layer from the proposed network encourages learning common features from diagnosing AMD and DR classes. This result also demonstrates that gradual changes in lesion features from different disease levels within classes provide auxiliary information for the model training. 

The same experiments have been repeated for the synergic adversarial label (Sec.~\ref{sec: al}) and multi-task learning (Sec.~\ref{sec: ml}) setting for the purpose of ablation study.  We find that the performance of our model will be improved even if only temperature scaling is added. This shows that for the grading of disease,  adding temperature scaling to prevent the label to be over-confident can make the model fully learn the interrelationship within the class, such as the gradual appearance, change or diffusion of pathological regions. Similarly, with the influence of synergic adversarial label (Sec.~\ref{sec: al}) and multi-task learning (Sec.~\ref{sec: ml}), the model can learn some information between DR and AMD such as common features. The results also prove the assumption we proposed in Sec.~\ref{sec: ts}.

In Table.~\ref{table_t} and Fig. \ref{5}, we vary the value of $T$ and explore how temperature scaling parameter $T$ affects the overall performance and reliability. The best performance can be achieved when $T=8$ for DR and $T=9$ for AMD. We get the highest improvement on DR and AMD accuracy, which is $3.42\%$, $3.5\%$ respectively. Although different values of $T$ result in various score, adding temperature scaling always benefits the baseline model. 

The results in Table.~\ref{table_v} provide an intuitive view of how temperature scaling actually affects our labels used for training. 
We calculate the average maximum and variance of synergic adversarial label. After using temperature scaling, both the maximum and variance have been significantly reduced. In this way, the training process can be more dependent on the inter-relationship reflected by labels within the class, rather than being affected only by the maximum value.

\begin{figure*}[!t]
	
	\centering
	
	\includegraphics[scale=.5]{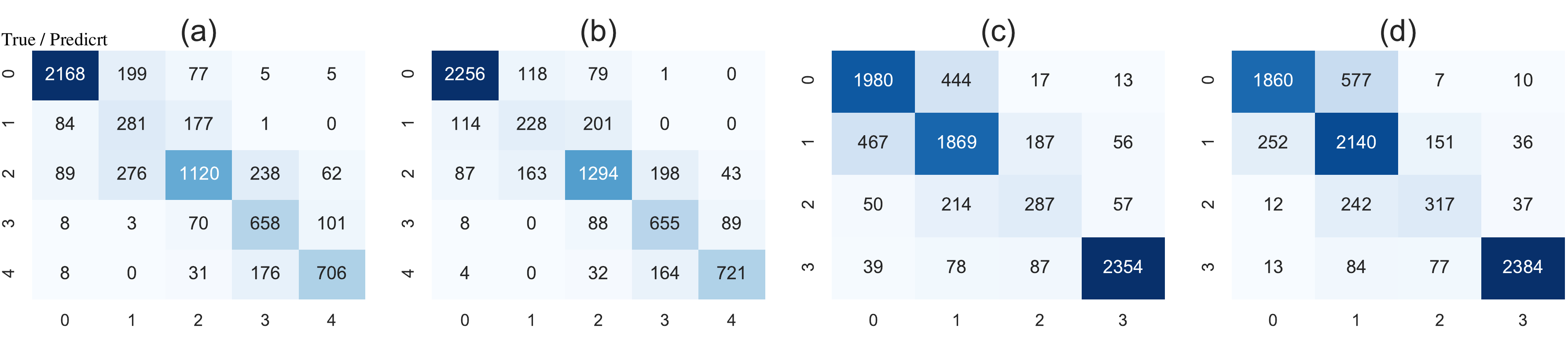}

	\caption{The confusion matrix. (a) Baseline model for DR; (b) SALL for DR; (c) baseline model for AMD; (d) SALL for AMD. The numbers in the axis denote the disease levels.}
	
	\label{matrix}
	
\end{figure*}

We output a confusion matrix of the baseline model and SALL (T = 8) in Fig. \ref{matrix} and observe the impact of our methods on classification results in more details. In the grading problems, we need to strictly consider the differentiation of normal, mild and severe diseases, especially normal and severe diseases, because this often involves whether to make a referral. It can be clearly noted that after using the SALL method, although the performance of some categories (such as NPDRI and NPDRII) will decline since the features of these categories are difficult to distinguish in clinical diagnosis. However, the cases that are misclassified to the most severe levels (CNV and PDR) are significantly reduced. At the same time, misclassification to normal also decreased. In other words, our model is more able to distinguish between normal and severe lesions. This makes grading more reliable and effective than just classification.

According to the above results, we summaries the following sub-conclusions: (1) When temperature scaling is not being employed, using synergic adversarial label augmentation solely directly on single-model training put bad influence on model training and leads to worse performance (The upper half of experiment-I). (2) Multi-task combined with either synergic adversarial label augmentation or temperature scaling achieves better performance (experiment-I). (3) While using temperature scaling, synergic adversarial label augmentation and multi-task together can achieve the best overall performance (experiment-II).

\subsection{Reliability Analysis with Calibration}
\begin{table*}[t]
	\centering
	\small
	\caption{Experiment-II: The ablation study results of different values of T.}
	\begin{tabular}{ccccccccc}
		\hline
		T & SA-label & Multi-task & DR acc & AMD acc & DR con & AMD con & DR err & AMD err \\ \hline
		4 & yes & yes & 77.73\% & 81.13\% & 86.39\% & 86.70\% & 66.29\% & 70.44\% \\
		5 & yes & yes & 78.30\% & 80.56\% & 85.32\% & 85.31\% & \textbf{65.26\%} & \textbf{68.10\%} \\
		6 & yes & yes & 78.53\% & 81.23\% & 85.57\% & 85.88\% & 65.33\% & 69.04\% \\
		7 & yes & yes & 78.39\% & 81.39\% & 85.36\% & 85.82\% & 65.48\% & 69.40\% \\
		8 & yes & yes & \textbf{78.77\%} & 81.73\% & \textbf{87.79\%} & \textbf{88.29\%} & 69.05\% & 73.49\% \\
		9 & yes & yes & 78.01\% & \textbf{82.02\%} & 87.53\% & 88.21\% & 67.56\% & 72.86\% \\
		10 & yes & yes & 77.97\% & 81.47\% & 86.72\% & 87.20\% & 67.63\% & 71.06\% \\
		20 & yes & yes & 78.40\% & 81.52\% & 84.75\% & 85.76\% & 66.66\% & 72.11\% \\
		50 & yes & yes & 77.81\% & 80.95\% & 86.74\% & 87.08\% & 69.94\% & 74.51\% \\ \hline
	\end{tabular}

	\label{table_t}
\end{table*}

\begin{figure*}[!t]
	
	\centering
	
	\includegraphics[scale=.5]{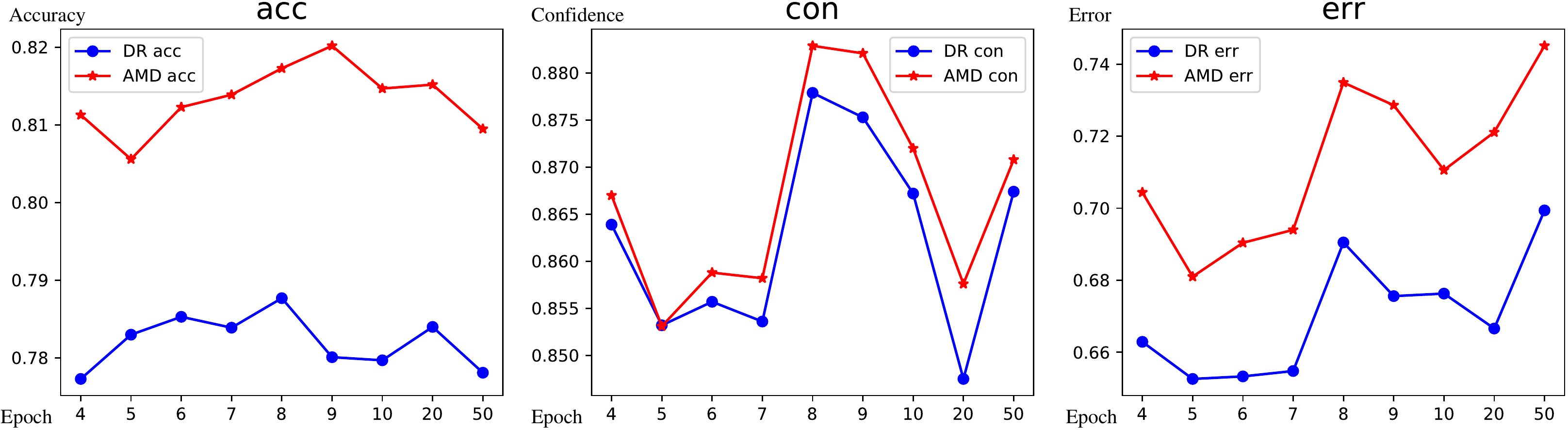}

	\caption{The results of different values of T on our model. We expected to improve the accuracy and confidence but reduce the error of it. Appropriate T values can help us improve the model by changing the confidence of label, but if the T value is too high, we will lose the distinction of information within and between classes and the performance of the model will also decline.}
	
	\label{5}
	
\end{figure*}


\begin{table}[t]
	\centering
	\small
	\caption{The maximum and variance of synergic adversarial label}
	\begin{tabular}{ccc}
		\hline
		\textbf{T} & \textbf{Max} & \textbf{Var} \\ \hline
		1 & 0.7478 & 0.0960 \\
		3 & 0.6915 & 0.0633 \\ \hline
	\end{tabular}
	
	\label{table_v}
\end{table}

As we discussed in Sec.~\ref{sec: ts}, learning very deep models with synergic adversarial label directly may produce overconfident predictions and affect model's training. In this part, we add temperature scaling to our model and analysis its effectiveness on training and calibrating the model. 
Following~\cite{Guo2017On}, for the first time, we formally introduce and formulate a few metrics to quantify the model reliability. 
We define of model confidence and error in Table.~\ref{table_1}, Table.~\ref{table_t} and Table.~\ref{table_ratio} as follows:
\begin{align*}\label{...}
Conf = Avg(\sum max(\bar{Y_{i}})) \mid argmax(\bar{Y_{i}}) = argmax(Y_{i})\\
Err = Avg(\sum max(\bar{Y_{i}})) \mid argmax(\bar{Y_{i}}) \neq argmax(Y_{i})
\end{align*}

where $\bar{Y_{i}}$ is the output from the Softmax layer, and $Y_{i}$ is the ground truth label. 
From Table.~\ref{table_t} and Fig.~\ref{5}, we find that our temperature scaling method improves average confidence probability for correctly predicted data from 85.74\% to 87.79\%, 86.06\% to 88.29\% respectively. Meanwhile, as for incorrectly predicted data, our method is able to reduce average error from 65.39\% to 65.26\%, 78.81\% to 68.10\% respectively. The need to pay attention to is that the error of AMD reduces a lot. Firstly, because of the influence of the training data(there may be some noise), although the training results of AMD have high accuracy, but also with high error. Secondly, by comparison with Table.~\ref{table_1}, we found that the decrease in errors was mainly due to the multi-task learning. It is the pathological information provided by DR in the weights sharing layers that helps AMD to reduce its errors to the same level as DR.

\begin{figure}[!t]
	\centering
	\includegraphics[scale=.35]{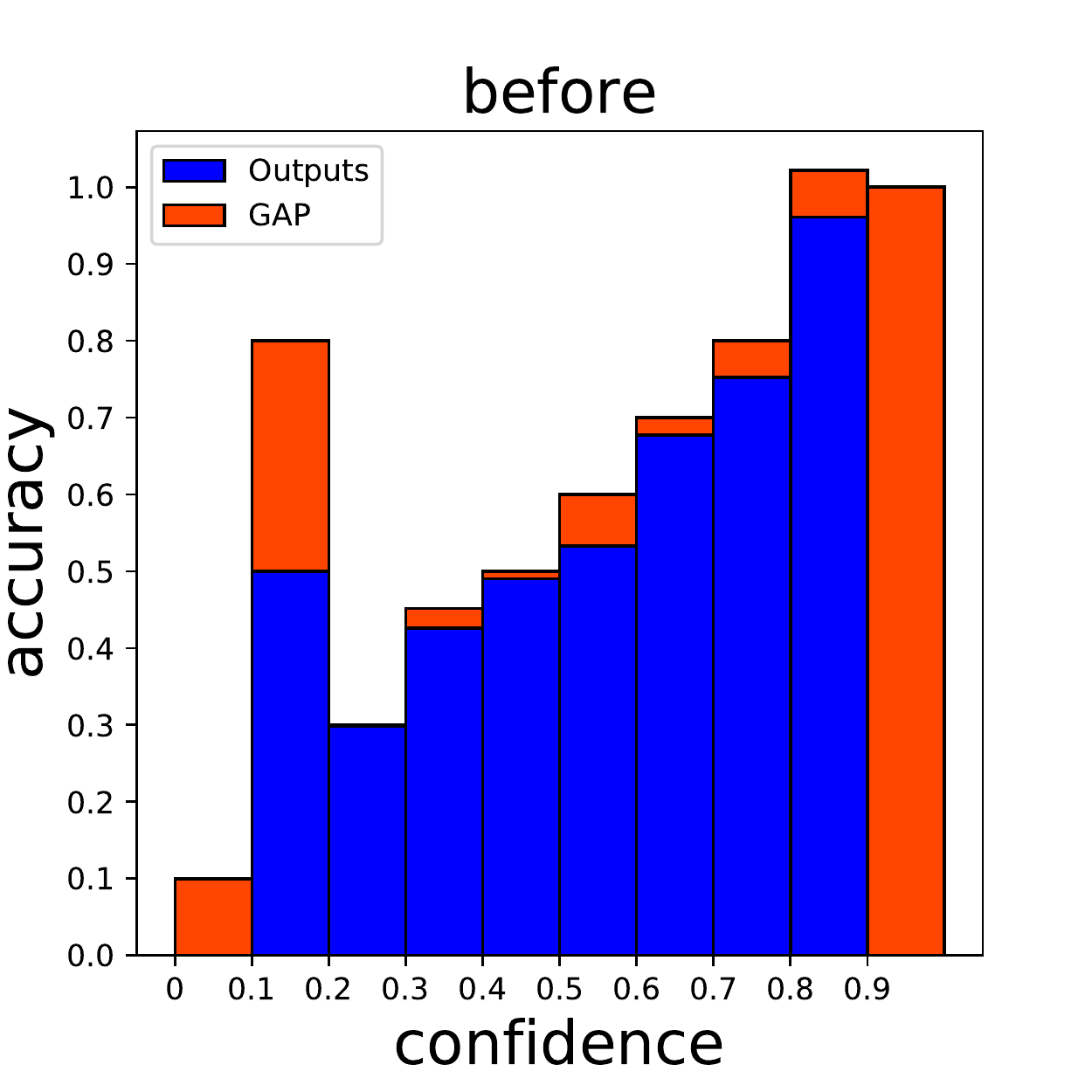}
	\includegraphics[scale=.35]{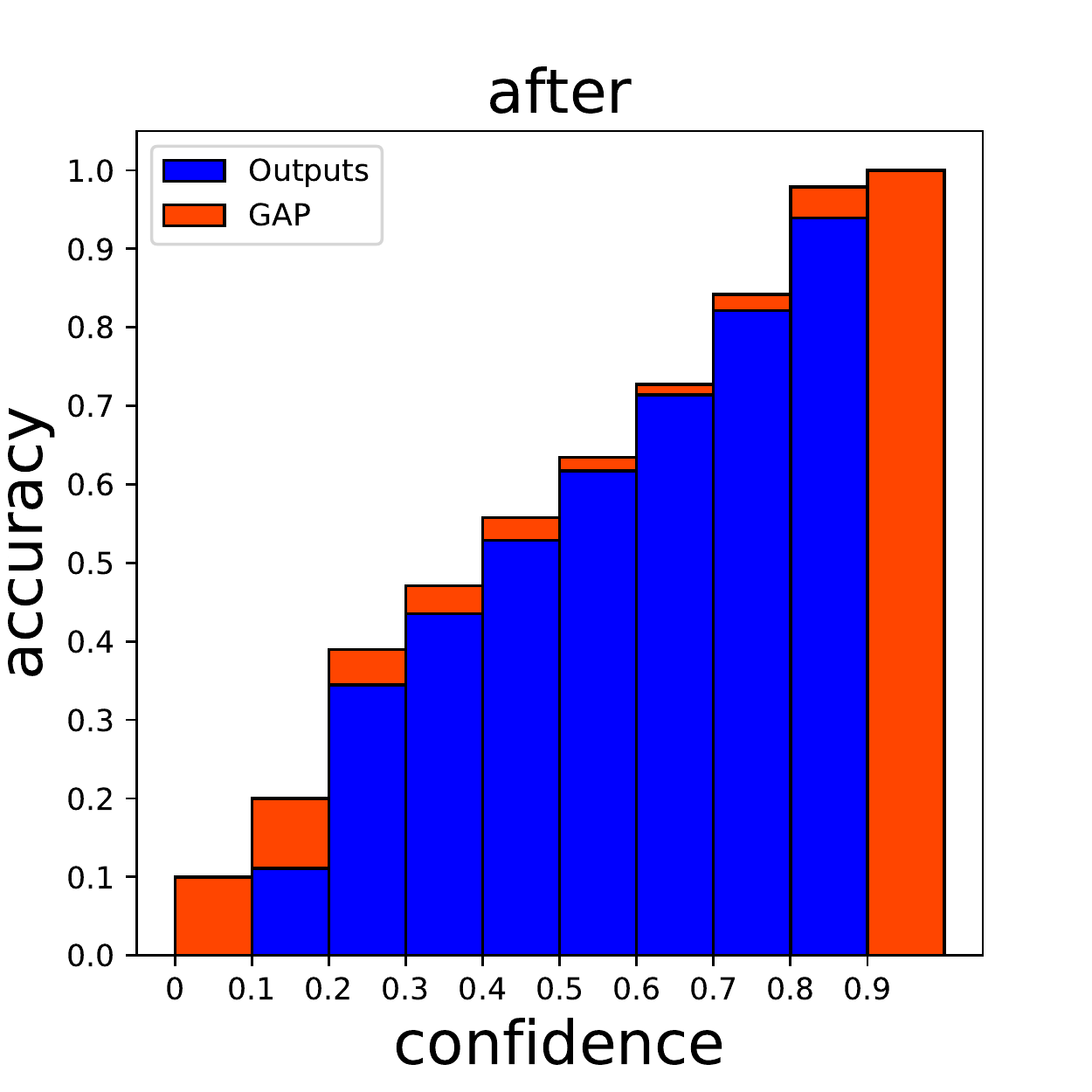}
	\caption{Reliability diagrams for single-task model and our proposed model. The red part in the figure represents the difference between the real value and the theoretical value. By comparison, it is found that the reliability of the model has been greatly improved based on our method. }
	\label{diagrams}
\end{figure}

To further verify the correlation between accuracy and confidence and reliability of the neural network model, we introduce reliability diagrams to help us evaluate whether the reliability of the model has been improved using temperature scaling. 
Please see details about reliability diagrams in~\cite{Guo2017On}. The ideal \emph{perfect calibration} is defined as:
\begin{equation}\label{...}
\mathbb{P}(\bar{Y}=Y|\bar{P}=p) = p, \; \forall p\in [0,1]
\end{equation}
where the probability is over the joint distribution. However in most practical settings, achieving perfect calibration is nearly impossible.
We plot reliability diagrams at $T = 3$ in Fig.~\ref{diagrams}. Fig. \ref{diagrams} are a visual representation of model calibration. These diagrams plot expected sample accuracy as a function of confidence. If the model is perfectly calibrated – i.e. if (11) holds – then the diagram should plot the identity function. Any deviation from a perfect diagonal represents miscalibration.
As can be seen that, our model has been well-calibrated by using temperature scaling through comparing the red chart in two diagrams. 
\begin{figure*}[!t]
	\centering
	\includegraphics[scale=.45]{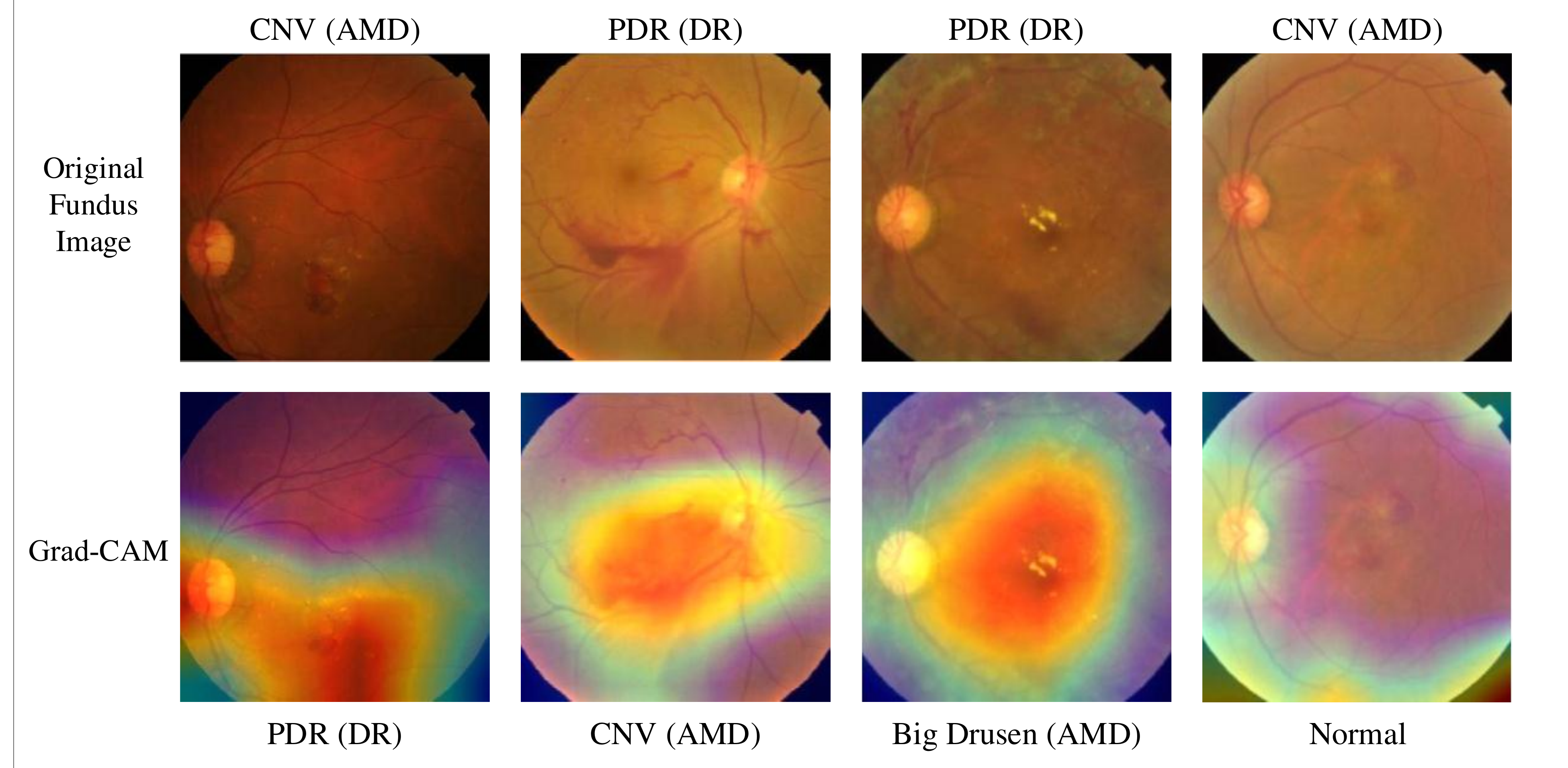}
	\caption{Gradient-weighted Class Activation Mapping (Grad-CAM) technique allows the classification-trained CNN to localize class-specific image regions. For example,  although CNV does not belong to any class in DR, we can still use the DR model to extract some information from AMD image. }
	\label{cam}
\end{figure*}

\subsection{Interpretability Analysis}
In the previous experiments, we have demonstrated the benefits of introducing implicit information into model training brought by synergic adversarial label. In this section, we try to perform visual analysis on how those extra synergic adversarial labels are helping and whether there are any direct connections between learned image features and DR or AMD pathologies. To obtain interpretability of our method, we introduce Gradient-weighted Class Activation Mapping (Grad-CAM)~\cite{selvaraju2017grad}. Grad-cam uses a specific class as input to calculate the weights of the feature map, and then overlay the original image to obtain a heat map. More details please refer to our citation. As shown in Fig.~\ref{cam}, we show that the Grad-CAM output of four different lesion pictures from the single-task model. The first row indicates original images with its ground truth labels while the second-row pictures illustrate the Grad-CAM output with opposite tasks (e.g. DR images into AMD network branch). 
The highlighted regions show where the model is focused on to diagnose the images. 

The leftmost image is a CNV image in AMD category. In DR single-task model, it is diagnosed as PDR. Through the visualization of CAM, there shows a lot of hemorrhagic points, which are common but important features for both CNV and NPDRIII. 
In pairs from the second column, we feed PDR images with haemorrhage parts into an AMD model, the final prediction from AMD model is judged as CNV. This observation can be explained with the same reason as the previous one. 
In the third column, the results from the AMD model seems not surprising since big drusen are visually similar to the exudation. 
In the last column, the DR model classifies it as normal because there are no obvious lesions features are being found.
According to the above visualisation results, those findings in the opposite model validates our assumption for our proposed method, which DR samples may provide useful features to learn a better model for AMD classifier because some features are common in both disease groups.

\subsection{DR and AMD Data Ratio Analysis}
\label{sec: dra}
The above experiments have proved that the accuracy of DR and AMD can be significantly improved by using our proposed method through synergic adversarial label augmentation and multi-task learning. Here we did an extra study to analysis the performance changes with respect to DR and AMD synergic adversarial training data ratio. We gradually change the synergic adversarial samples ratio under the best setting we achieved in Table.~\ref{table_1} with $T=3$, synergic adversarial label and multi-task learning.
The results are shown in Table.~\ref{table_ratio}. We find that balanced DR and AMD proportion does not result in optimal performance. By adjusting the synergic adversarial ratio as a hyper-parameter, we can further improve the performance of the model by $3.52\%$ and $2.32\%$ on DR and AMD task respectively. 

Synergic adversarial label has increased the amount of training data, and multi-task learning maximizes the effect of it. What is more,  as we discussed in Sec.~\ref{sec: ts}, we added temperature scaling to help the model better distinguish useful and useless information. In this section, we try to achieve the same effect by changing the data ratio. For example, for DR training tasks, we can regard the irrelevant features brought by AMD as a kind of noise, which can be reduced by change the amount of AMD data appropriately. However, if the amount of AMD data is too small, the useful information we need will be greatly reduced too, and the performance of DR branch model will also be reduced.

\begin{table}[]
	\centering
	\caption{The performance of model with train data on different data ratio.}
	\begin{tabular}{lllll}
		\hline
		\multicolumn{1}{c}{DR} & \multicolumn{1}{c}{AMD} & \multicolumn{1}{c}{DR acc} & DR con & DR err \\ \hline
		\multicolumn{1}{c}{28154} & \multicolumn{1}{c}{400} & \multicolumn{1}{c}{77.92\%} & 87.51\% & 67.79\% \\
		\multicolumn{1}{c}{28154} & \multicolumn{1}{c}{2800} & \multicolumn{1}{c}{78.07\%} & 86.89\% & 67.86\% \\
		\multicolumn{1}{c}{28154} & \multicolumn{1}{c}{13101} & \multicolumn{1}{c}{77.99\%} & 87.54\% & 66.91\% \\
		\multicolumn{1}{c}{28154} & \multicolumn{1}{c}{18844} & \multicolumn{1}{c}{\textbf{78.91\%}} & \textbf{88.55\%} & 65.79\% \\
		\multicolumn{1}{c}{28154} & \multicolumn{1}{c}{28228} & \multicolumn{1}{c}{78.01\%} & 85.68\% & \textbf{63.10}\% \\
		\hline
		&  &  &  &  \\ \hline
		\multicolumn{1}{c}{DR} & AMD & AMD acc & AMD con & AMD err \\ \hline
		\multicolumn{1}{c}{500} & 28228 & 80.70\% & 87.71\% & 71.37\% \\
		\multicolumn{1}{c}{3000} & 28228 & 80.85\% & \textbf{87.77\%} & 71.49\% \\
		15533 & 28228 & \textbf{81.47\%} & 87.44\% & 70.43\% \\
		19620 & 28228 & 80.08\% & 87.21\% & 70.96\% \\
		28154 & 28228 & 81.25\% & 86.15\% & \textbf{65.75\%} \\ \hline
	\end{tabular}

	\label{table_ratio}
\end{table}

\subsection{3 muti-task}
\begin{figure}[!t]
	\centering
	\includegraphics[width=6cm]{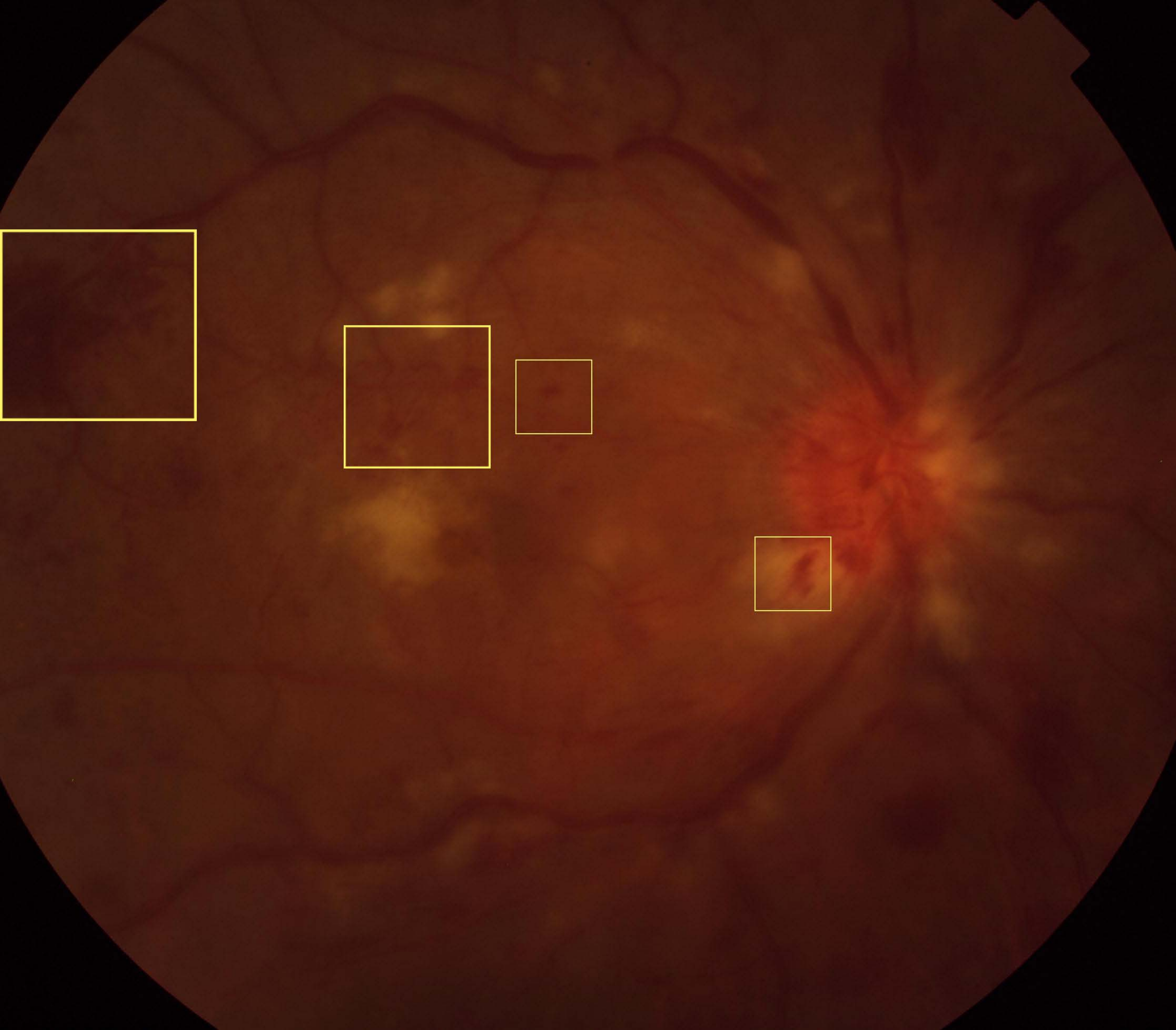}
	\caption{The lesion features in arteriosclerosis.}
	\label{arter}
\end{figure}

\begin{table}[!t]
	\centering
	\caption{The amount of newly added datasets after augmentation.}
	\begin{tabular}{cccc}
		\hline
		& \textbf{Train} & \textbf{Validation} & \textbf{Test} \\ \hline
		Normal           & 7359           & 2454                & 2454          \\
		Arteriosclerosis & 2762           & 921                 & 921           \\ \hline
	\end{tabular}
	
	\label{arter}
\end{table}

\begin{table*}[!t]
	\centering
	\caption{The performance of 3 sub-task model.}
	\begin{tabular}{p{70pt}p{28pt}p{28pt}p{28pt}p{28pt}p{28pt}p{28pt}p{28pt}p{28pt}p{28pt}}
		\hline
		
		& DR acc           & AMD acc          & Arter acc        & DR con           & AMD con          & Arter con        & DR err           & AMD err          & Arter err        \\ \hline
		Single Model & 75.39\%          & 79.15\%          & \textbf{81.95\%} & 85.74\%          & 86.06\%          & 83.14\%          & 65.39\%          & 78.81\%          & 62.45\%          \\
		2 sub-task & \textbf{78.77\%} & \textbf{81.73\%} & NA               & \textbf{87.79\%} & \textbf{88.29\%} & NA               & 69.05\%          & 73.49\%          & NA               \\
		3 sub-task & 78.23\%          & 81.07\%          & 81.87\%          & 85.08\%          & 86.03\%          & \textbf{85.79\%} & \textbf{63.92\%} & \textbf{67.88\%} & \textbf{61.82\%} \\ \hline
	\end{tabular}
	
	\label{3-task}
\end{table*}
In this section, we changed the subtask branch of our network structure to three and added the data for arteriosclerosis. In arteriosclerosis, degenerative changes occur in the walls of arteries that lead to thickening of arterial walls and narrowing of blood vessels and may give rise to complete occlusion (blockage) of a vessel. Blockage of retinal veins results in the bursting of small vessels, retinal swelling, and multiple haemorrhages scattered over the retina \cite{EyeDisease} as Fig.~\ref{arter} shown. Compared with DR and AMD, Arter only consists of one disease level and has less similarity with DR to AMD. Here, we aim to explore the changes with varying levels of similarity between sub-tasks.

The amount of newly added datasets is shown in Table.~\ref{arter} and the performance of 3 sub-task model is shown in Table.~\ref{3-task} where we set T=8. For comparison, we add the result of 2 sub-task from Table.~\ref{table_t}. 

From the Table.~\ref{3-task}, we can find that the performance of DR and AMD is still improved compared with baseline model, but the change is not obvious compared with 2 sub-task, and we can notice that the error is significantly reduced. However, for arteriosclerosis, there is no obvious change except confidence and error, which benefit from temperature calibration. 

In order to explain the above results, we found that the newly added data of arteriosclerosis compared with the data of DR and AMD, although we have achieved the data augmentation, there is still imbalance (the data of DR and AMD are at least twice of that of arteriosclerosis). As we discussed in Sec.~\ref{sec: dra}, we believe that the data ratio is an important factor in multi-task classification since there is a trade-off between sub-tasks with different levels of similarity, and too much irrelevant information may do harm to the performance. Therefore, we have reduced the number of samples for each class, in order to make a balance for them, and the results are shown in Table.~\ref{3-task-2}.

Although the performance on arteriosclerosis has not been greatly improved, we are surprised to find that when we limit 3000 images to each class, performance on DR reaches the highest level in this paper based on our proposed method, and when there are only 1500 images for each class, the classification accuracy of DR is also higher than that of the baseline model (full datasets) and achieves 4.35\% improvement. This indicates that our method still has strong applicability for a small amount of data. We need to emphasize that our method does not require a great effect for every sub-task. If the performance of one or two of the three sub-task is improved but the effect of the other sub-task is decreased, then we can still think our method is effective.

\begin{table}[t]
	\centering
	\caption{The performance for 3 sub-task model with different amount of datasets for each class.}
	\begin{tabular}{p{100pt}ccc}
		\hline
		Amount/Class & DR acc                               & AMD acc          & Arter acc        \\ \hline
		1500 + Single Model        & 70.32\%                              & 70.61\%          & 80.89\%          \\
		1500 + ours    & \textbf{76.23\%}                     & \textbf{74.30\%} & \textbf{81.11\%} \\
		3000 + Single Model        & 74.72\%                              & 75.58\%          & \textbf{81.64\%} \\
		3000 + ours    & \multicolumn{1}{l}{\textbf{79.74\%}} & \textbf{76.17\%} & 81.12\%          \\ \hline
	\end{tabular}
	
	\label{3-task-2}
\end{table}

\section{Conclusion}

In this paper, we propose a novel learning method to improve the performance of retinal imaging diagnosis by using the knowledge from different datasets containing retinal diseases with high similarity. Our main technical innovations include two main parts, synergic adversarial labels generation and multi-task learning. In the future, we will focus on two things, first using more relevant retinal disease images and explore the further relationship between those labels to enlarge the scale of our proposed method; secondly, we will investigate the possibility of generalisation of this method into general image classification domain.  

\bibliographystyle{IEEEtran.bst}
\bibliography{refs.bib}
\end{document}